\documentclass[letterpaper, 10 pt, conference]{ieeeconf}  

\usepackage{booktabs}       
\usepackage{caption}
\usepackage{amsfonts}       
\usepackage{nicefrac}       
\usepackage{microtype}      
\usepackage{xcolor}         
\usepackage{graphicx}
\usepackage{multirow}
\usepackage{wrapfig}
\usepackage[hang,flushmargin]{footmisc}

\usepackage{comment}
\usepackage{amsmath,amsfonts,bm}

\IEEEoverridecommandlockouts                              

\title{\LARGE \bf
Zero-Shot Open-Vocabulary Tracking with \\Large  Pre-Trained Models
}

\author{Wen-Hsuan Chu$^{1}$, Adam W. Harley$^{2}$, Pavel Tokmakov$^{3}$, Achal Dave$^{3}$, Leonidas Guibas$^{2}$, Katerina Fragkiadaki$^{1}$\vspace{-1em}%
\thanks{$^{1}$Carnegie Mellon University {\tt\small <wenhsuac,katef>@cs.cmu.edu}, $^{2}$Stanford University {\tt\small <aharley,guibas>@cs.stanford.edu}, $^{3}$Toyota Research Institute {\tt\small <pavel.tokmakov,achal.dave>@tr}
{\tt\small i.global}}
}

\begin{document}

\maketitle

\newcommand{\model}{OVTracktor}
\newcommand{\smallsec}[1]{\vspace{0.0em}\textbf{#1}}

\let\ab\allowbreak

\newcommand{\etal}{\text{et al.}}

\def\fig#1{Fig.~\ref{fig:#1}}
\def\imw#1#2{\includegraphics[width=#2\linewidth]{#1.png}}
\def\imwjpg#1#2{\includegraphics[width=#2\linewidth]{#1.jpg}}
\def\imh#1#2{\includegraphics[height=#2\textheight]{#1.png}}
\def\imhjpg#1#2{\includegraphics[height=#2\textheight]{#1.jpg}}
\def\imwh#1#2#3{\includegraphics[width=#2\linewidth,height=#3\textheight]{#1.png}}
\def\imwhjpg#1#2#3{\includegraphics[width=#2\linewidth,height=#3\textheight]{#1.jpg}}
\newcommand{\tb}[3]{\setlength{\tabcolsep}{#2mm}\begin{tabular}{#1}#3\end{tabular}}
\newcommand{\ol}[3]{\begin{#1}[leftmargin=*,topsep=1pt]\setlength{\itemsep}{#2pt}#3\end{#1}}

\newcommand{\xmark}{\textcolor{red}{\ding{55}}}
\newcommand{\greencheck}{\textcolor{green}{$\checkmark$}}
\newcommand{\bluehighlight}[1]{{\color{blue}#1}}
\newcommand{\greenhighlight}[1]{{\color{teal}#1}}
\newcommand{\redhighlight}[1]{{\color{red}#1}}
\newcommand{\bzero}{\mathbf{0}}
\newcommand{\bone}{\mathbf{1}}
\newcommand{\cat}{\ell}
\newcommand{\cond}{\textbf{c}}
\newcommand{\bluebf}[1]{{\color{blue} \textbf{#1}}}
\newcommand{\greenbf}[1]{{\color{teal} \textbf{#1}}}
\newcommand{\redbf}[1]{{\color{red} \textbf{#1}}}
\newcommand{\fixme}[1]{{\color{red} \textbf{#1}}}
\newcommand{\fixed}[1]{{\color{blue} #1}}

\newcommand{\algorithmName}{Test-time adaptation with Diffusion Models\xspace}
\newcommand{\algorithmShort}{Diff-TTA\xspace}
\thispagestyle{empty}
\pagestyle{empty}


\begin{abstract}
Object tracking is central to robot perception and scene understanding. Tracking-by-detection has long been a dominant paradigm for object tracking of specific object categories~\cite{citeulike:3504612,baseline1}. Recently, large-scale pre-trained models have shown promising advances in detecting and segmenting objects and parts in 2D static images in the wild. This begs the question: can we re-purpose these large-scale pre-trained static image models for open-vocabulary video tracking? In this paper, we re-purpose an open-vocabulary detector~\cite{zhou2022detecting}, segmenter~\cite{kirillov2023segment}, and dense optical flow estimator~\cite{GMFlow}, into a model that tracks and segments objects of any category in 2D videos.
Our method predicts object and part tracks with associated language descriptions in monocular videos,
rebuilding the pipeline of Tractor \cite{bergmann2019tracking} with modern large pre-trained models for static image detection and segmentation: we detect open-vocabulary object instances and propagate their boxes from frame to frame using a flow-based motion model, refine the propagated boxes with the box regression module of the visual detector, and prompt an open-world segmenter with the refined box to segment the objects. We decide the termination of an object track based on the objectness score of the propagated boxes, as well as forward-backward optical flow consistency. We re-identify objects across occlusions using deep feature matching. We show that our model achieves strong performance on multiple established video object segmentation and tracking benchmarks~\cite{Pont-TusetPCASG17, XuYFYYLPCH18, uvo, athar2023burst}, and can produce reasonable tracks in manipulation data~\cite{vecerik2023robotap}. In particular, our model outperforms previous state-of-the-art in UVO and BURST, benchmarks for open-world object tracking and segmentation, despite never being explicitly trained for tracking. We hope that our approach can serve as a simple and extensible framework for future research. The project page can be found \href{https://wenhsuanchu.github.io/ovtracktor/}{\underline{here}}.
\end{abstract}

\section{Introduction}
\label{intro}
We are interested in the problem of tracking arbitrary objects in video. A reasonable strategy for this task, which has dominated the area for multiple years, is ``tracking by detection''~\cite{tbd}. Tracking by detection splits the task into two independent problems: (1) detect objects frame-by-frame, and (2) associate detection responses across frames. Because of its two-stage split, tracking-by-detection is mainly propelled forward by advances in detection. Notably, Tracktor~\cite{bergmann2019tracking} 
used person detections from a Faster R-CNN~\cite{FasterRCNN}, propagated boxes with a simple motion model, and refined these boxes using the Faster-RCNN's box regression head. This yielded a tracker composed entirely of static image neural modules, far simpler than its contemporary methods while matching or exceeding their accuracy. Recent methods for visual tracking build upon transformer architectures, where feature vectors represent tracked objects, and these are re-contextualized in each frame by attending to the pixel features and are used to predict per-frame bounding boxes~\cite{trackformer,athar2022hodor,DBLP:journals/corr/abs-2012-15460}.
These methods are trained on annotated video data and do not capitalize on pre-trained static image detectors beyond the pre-training of their feature backbones \cite{trackformer}.  

Meanwhile, 2D image object detection has been recently revolutionized with open-world image detectors, jointly trained for referential grounding and category grounding of thousands of object categories \cite{zhou2022detecting,liu2023grounding,zhang2022glipv2} across millions of images.  
Can we capitalize on this and make practical progress for tracking-by-detection, by re-visiting the Tracktor paradigm~\cite{bergmann2019tracking} with these updated components? 
In other words, can we re-purpose large pre-trained image models into a \textit{zero-shot} open-vocabulary tracker, without ever fine-tuning on video data?

We propose a simple and extensible framework for exploring this question, which does not introduce significant advancements or innovative approaches. 
We use an open-vocabulary detector to find objects as they appear~\cite{zhou2022detecting}, obtain their masks using an off-the-shelf segmenter~\cite{kirillov2023segment}, propagate the boxes to the next frame using a motion transformation computed from optical flow~\cite{GMFlow}, refine the boxes using the detector's bounding box regression module, and segment the box interiors using an off-the-shelf segmenter~\cite{kirillov2023segment}. 
We handle ambiguity in per-frame segmentations by selecting the segmentations with the highest temporal consistency. Finally, we revise the bounding boxes using the segmentation results.

We test our method on multiple established video object segmentation and tracking benchmarks: UVO \cite{uvo} and BURST \cite{athar2023burst}, as well as traditional VOS benchmarks like DAVIS \cite{Pont-TusetPCASG17} and YoutubeVOS \cite{XuYFYYLPCH18}. In open-world datasets like UVO and BURST, our method outperforms the previous state-of-the-art and achieves competitive performance in DAVIS and YoutubeVOS when evaluating VOS baselines using detected first-frame masks. Our tracker can also provide reasonable object tracks in RoboTAP \cite{vecerik2023robotap}, a manipulation-based dataset. Our method also provides a natural-language interface for tracking, where a user may describe the tracking target in words, and the model delivers the corresponding frame-by-frame segmentation. Given that our approach can improve as new pre-trained models are swapped in, we hope that our approach will serve as a simple yet extensible framework for future work.

\section{Related Work}
\label{related}

\begin{figure*}[!t]
\centering{}
\includegraphics[width=0.95\linewidth]{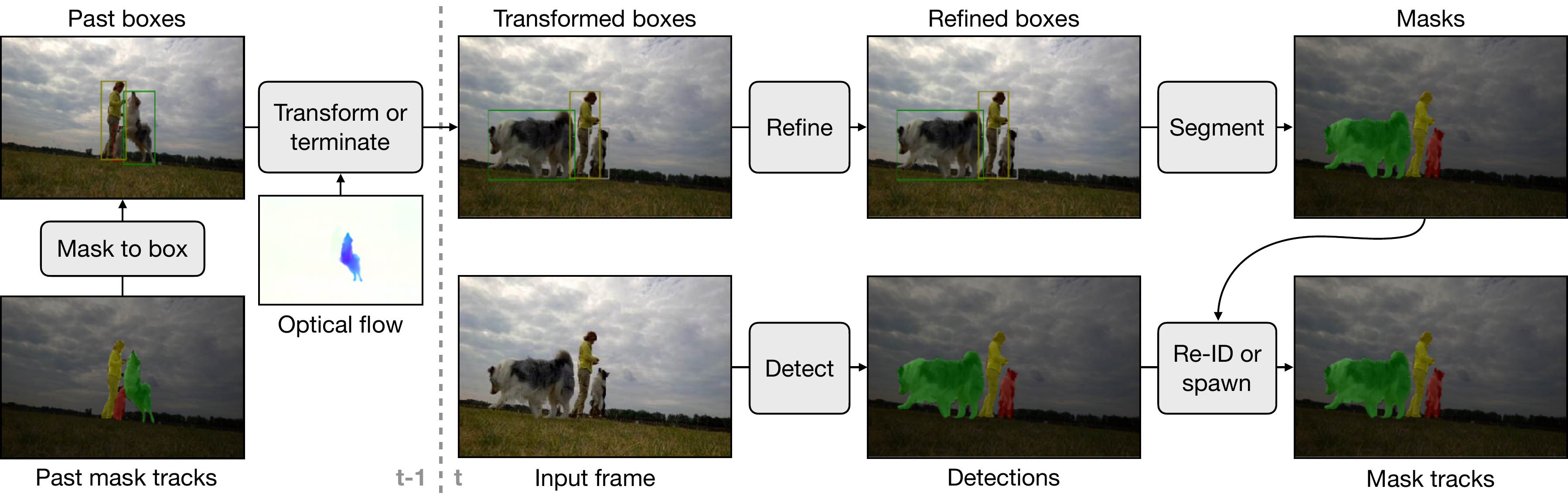}
\caption{\textbf{Architecture of \model{}}. An open-vocabulary detector detects objects and an open-world segmenter segments their masks. We propagate boxes to the next frame or decide their termination using an optical flow-based motion model. The propagated boxes are refined with the detector's box regression module. Refined boxes are used to prompt the segmenter in the next frame. The detections and their associated appearance features from the next frame are used to determine whether new tracks should be spawned or merged with previously terminated tracks.}
  \label{fig:method}
\end{figure*}

\smallsec{Tracking by detection.}
Many modern object tracking approaches rely heavily on accurate per-frame \textit{detectors} \cite{zhou2020tracking,bergmann2019tracking,bewley2016simple,wojke2017simple,tang2017multiple,zhang2021fairmot}.
These approaches show that simple post-processing of per-frame detection can lead to strong tracking approaches for a closed (usually small) set of objects. In particular, CenterTrack~\cite{zhou2020tracking} also showed that a tracker can be obtained simply from training on (augmented) static images, by modeling humans and vehicles as points. However, in open-world settings, this is less feasible, as objects may overlap and have different sizes (e.g. the upper half body of a person and their shirt), making points an ambiguous descriptor for open-world tracking.
Our work is most closely related to Tracktor~\cite{bergmann2019tracking}, which directly uses a Faster R-CNN \cite{FasterRCNN} person detector to build an accurate person tracker.
Our work builds on this method, extending it to \textit{any} category, using a strong open-vocabulary detector \cite{zhou2022detecting} as the backbone.


\smallsec{Open vocabulary detection.}
Recent advances in open-vocabulary classification \cite{radford2021learning,jia2021scaling} have significantly improved open-vocabulary \textit{detectors}. 
Open-vocabulary, or zero-shot, detectors largely operate by using language models to generalize to unseen object classes.
Early approaches relied on using text embeddings from pre-trained language models, such as from BERT~\cite{devlin2018bert} or GLOVE~\cite{pennington2014glove}, as classifiers for object proposals \cite{zareian2021open,bansal2018zero,rahman2020improved}.
More recent work leverages text embeddings which are pre-trained to be aligned with vision embeddings~\cite{radford2021learning,jia2021scaling}, leading to significant improvements in accuracy~\cite{zhou2022detecting,gu2021open,minderer2022simple}. 
We show that this recent class of approaches can be directly generalized to open-vocabulary tracking, using Detic~\cite{zhou2022detecting} as a representative model.

\smallsec{Open world tracking.}
Object tracking has traditionally focused on a few categories, such as people and vehicles.
Very recently, the community has seen renewed efforts to generalize tracking to \textit{arbitrary} objects. 
Traditional approaches focused on \textit{motion}-based segmentation~\cite{shi1998motion,grundmann2010efficient,ochs2013segmentation,Fragkiadaki_2015_CVPR,bideau2018best}, leveraging motion as a cue to segment never-before-seen objects.
More recent approaches use open-world object proposal methods to detect objects per frame and link them together using a combination of temporal consistency, appearance, and motion cues~\cite{liu2022opening,li2022tracking,ovsep2018track}.
Our work extends this latter class of approaches to \textit{open-vocabulary} detectors.


\section{Method}
\label{method}
Our method builds upon existing open-vocabulary detectors~\cite{zhou2022detecting}, promptable general-purpose segmenters~\cite{kirillov2023segment}, 
and dense optical flow estimators~\cite{GMFlow}. 
We call our model Open-Vocabulary Multi-Object Tracker, or \model{} for short. 
Figure~\ref{fig:method} shows an illustration of our model architecture. 
Our model does not require any tracking-specific training.
In this section, we first introduce the modules that we rely on, then discuss how we combine them into \model{}. 
\subsection{Building Blocks}
\paragraph{Open-vocabulary object detector}
We use Detic~\cite{zhou2022detecting} with a Swin-B~\cite{LiuL00W0LG21} backbone as our open vocabulary object detector. Detic is a two-stage detector. In the first stage, it generates a large number of candidate boxes with a Region Proposal Network (RPN), similar to Faster-RCNN~\cite{FasterRCNN}. In the second stage, it spatially refines each box using a regression module and predicts an objectness score and category label for each. In addition, a category-agnostic mask prediction head is trained to segment the object in each predicted bounding box. Our model exploits Detic's ability to detect and label object boxes and also re-uses its bounding box refinement module during tracking. 

\paragraph{Promptable general-purpose segmenter}
For segmenting masks from object boxes, we rely on SAM~\cite{kirillov2023segment}, a recent interactive general-purpose segmenter, which produces a segmentation given box or point prompts that indicate the object of interest. SAM is a transformer-based~\cite{VaswaniSPUJGKP17} model with a large and high-resolution image encoder and a lightweight prompt-conditioned mask head. For each user prompt, SAM predicts multiple segmentation hypotheses. 

\paragraph{Optical flow estimation}
We estimate the motion transformation of an object box to propagate it from frame to frame. We use GMFlow~\cite{GMFlow}, which is a state-of-the-art optical flow method. GMFlow takes two consecutive frames as input, and produces a 2D pixel displacement map as output, using an architecture that computes a spatial argmax of feature correlations for each pixel, trained on large synthetic datasets. We use this flow map to estimate the motion of detected boxes and also rely on optical forward-backward flow cycle-consistency~\cite{pan2009recurrent} to estimate occlusion, in which case we terminate the track. 

\subsection{\model{}}
Given an RGB video as input, the goal of \model{} is to estimate mask trajectories for all objects in the video and estimate category labels for those objects. 
A mask trajectory for object $i$ is a sequence of image binary masks  $M_i = \{m_i^t\, |\, t \in \left[0, T\right]\}$, where $m_i^t \in \mathbf{R}^{W \times H}$, where $W \times H$ denote the width and height of the image frame, and $t$ is the frame index in time. Each object is associated with a category label, which we denote with $\cat_i$. We denote the set of all binary instance masks in frame $t$ as $M^t = \{m_0^t,\, m_1^t,\, \ldots \}$.
 
\paragraph{Detection}
We run the detector on every frame. 
Let $D^t$ denote the object detections and segmentations supplied by the detector at frame $t$. 
At $t=0$, our tracker initializes object masks $M^0$ from the set of Detic object detections $D^0$, thresholded at a  confidence threshold $\lambda_c=0.5$. 

\paragraph{Motion-driven box propagation} 
We propagate object boxes across consecutive frames using a 4-parameter box motion transformation that includes a box translation $(dx,dy)$ and width and height scaling $(s_x,s_y)$ using motion information obtained from an optical flow field of ~\cite{GMFlow}.
We filter the pixel displacement vectors that are forward-backward consistent \cite{conf/eccv/SundaramBK10}. A lenient criterion is used: we simply check if the forward-backward flows have segmentation consistency, in the sense that tracking forward and backward leads back to the original instance mask, instead of thresholding the forward-backward displacement from the origin.
We use the filtered pixels to compute a box motion transformation using least squares and use this to propagate the box forward. After this motion warp, the box is still axes-aligned, we do not consider object rotation or anisotropic scaling. The category label and instance ID of the box are maintained.

\paragraph{Object track termination}
We determine if an object track should be terminated due to occlusions by checking if the ratio of forward-backward flow consistent pixels is lower than a fixed ratio $\lambda_{flow}$ or if the object-ness score of the box is too low.

\paragraph{Object box refinement}
We refine the propagated (non-terminated) boxes at frame $t+1$ using Detic's bounding box regression module. This adjusts the bounding boxes according to objectness cues on that frame and gives us higher-quality box estimates in frame $t+1$. 

\paragraph{Temporally consistent object segmentation}
The bounding box estimates at frame $t+1$ are used to prompt SAM to segment the object's interior. SAM produces multiple segmentation mask candidates per box, to handle ambiguity regarding what to segment from the box's interior. Overall, we found a box prompt often unambiguously determines the object to segment (so all resulting masks will be identical), in contrast to a center object point prompt, which does not have information regarding the object's extent. To handle the cases where this ambiguity exists, we implement a form of temporal cycle consistency at the mask level.

SAM segments an object via iterative attention between an object query vector and pixel features, and a final inner product between the contextualized query vector and pixel feature vectors. The three segmentation hypotheses consider different object query initialization.  For each box $i$, we use the updated (contextualized) object query vector at frame $t+1$ to segment the object at frame $t$ via inner product with the pixel features from frame $t$; this results in a temporally corresponding mask $\hat{m}_i^{t}$. We select the SAM segmentation hypothesis at frame $t+1$ whose updated query vector-driven segmentation $\hat{m}_i^{t}$ has the highest Intersection over Union (IoU) with $m_i^{t}$. We then update the object boxes to tightly contain the resulting segmentation mask. 

\paragraph{Spawning new object tracks} At each frame, we need to take into account new objects that enter the scene or reappear after an occlusion. For each detection in $D^{t+1}$, we compute its IoU with all the masks in $M^t$. A new track is spawned if the IoU between the detection and all masks in $M^t$ is below some specified threshold $\lambda_{spawn}$. 

\paragraph{Track re-identification}
We use appearance feature matching to determine whether to merge a new track with an existing but terminated track. We store a small window of features before a track's termination and compare them with the features of the newly spawned tracks. Newly spawned tracks are considered for Re-ID until $T_{reid}$ time-steps have passed. We used the box features from Detic (before the box regression module) and normalized them along the channel dimension to obtain a small set of features that represent each instance. We then compute the inner product between normalized appearance features for any two tracks and merge them if their value is above a threshold $\lambda_{reid}$.

None of \model{}'s described components require any additional training. As detectors and segmenters improve, the components can be easily swapped out for potential improvements to the tracker.

\textbf{Implementation details.} 
During inference, we apply test time augmentation to video frames when running the detector by scaling and horizontally flipping the individual frames to obtain better object proposals, which we found to help improve the recall of detections, especially in the harder open-world datasets. \model{} has the following hyper-parameters $\lambda_c=0.5$ for thresholding detector's confidence, $\lambda_{flow}$ for deciding track termination due to occlusion, $\lambda_{spawn}$ for instantiating new objects non-overlapping with existing objects tracks, and  $\lambda_{reid}$ for merging temporally non-overlapping tracks during re-identification. We have found the model robust to the choice of these hyper-parameters, due to the nature of videos: an object suffices to be detected confidently only very sparsely in time, and our propagation method will propagate it forward. In evaluations, we used some videos in the training set to select a good set of hyperparameters for the individual datasets. As the memory consumption scales with the number of objects being tracked, we also put a hard limit $K$ on the limit of tracks that can co-exist at the same time. 

\section{Experiments}
\label{exps}
\begin{figure}[!t]
\includegraphics[width=\linewidth]{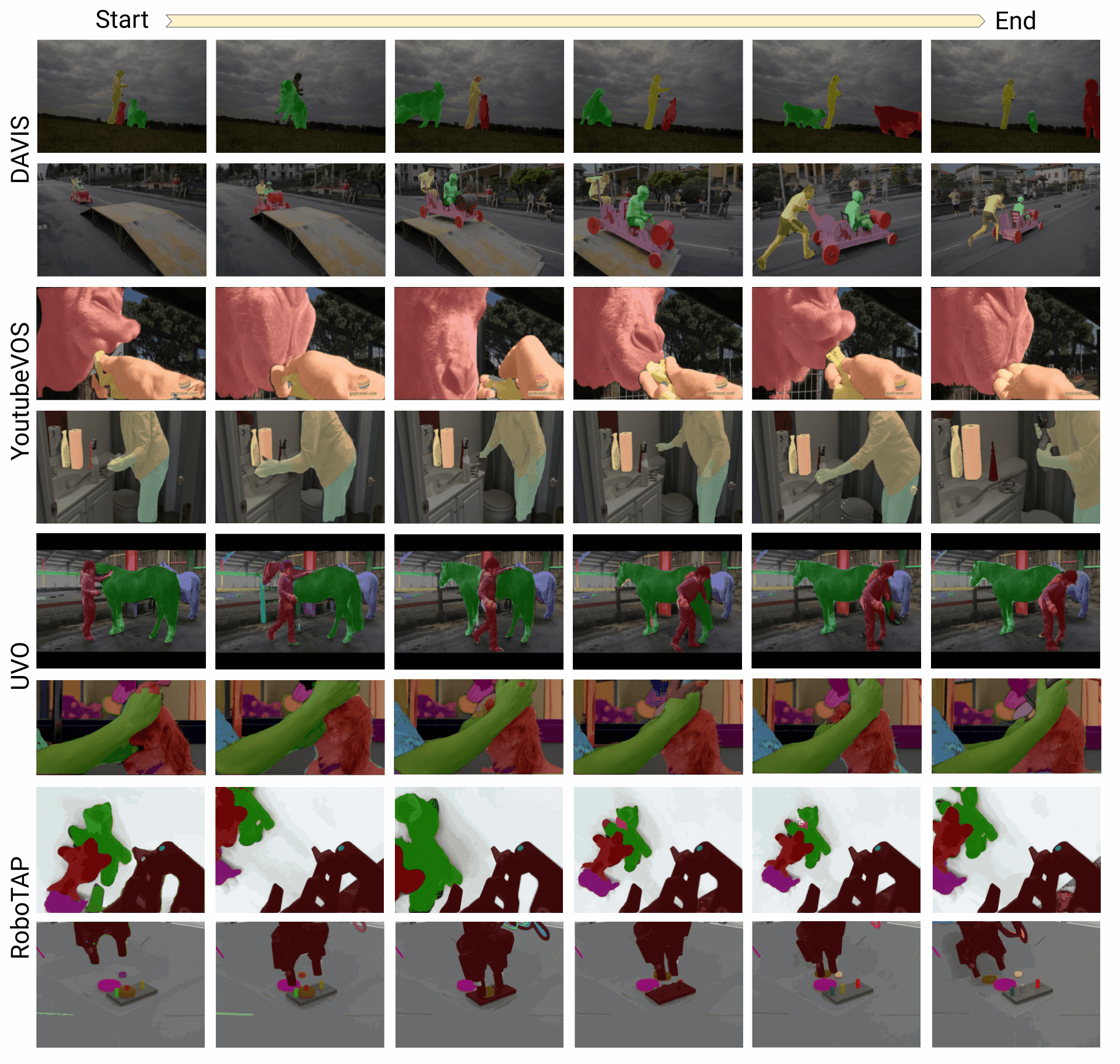}
\caption{\textbf{Qualitative object tracking results} in DAVIS'17, YoutubeVOS, UVO, and RoboTAP, with frames selected uniformly from the start (left) to the end (right) of a video. \model{} can detect and track objects consistently through time and can distinguish between similar-looking instances. 
}
  \label{fig:tracking_vis}
\end{figure}

We test \model{} in multi-object tracking and video object segmentation benchmarks of BURST~\cite{athar2023burst}, UVO~\cite{uvo}, DAVIS~\cite{Pont-TusetPCASG17}, and YoutubeVOS~\cite{XuYFYYLPCH18}, along with some ablation studies and analysis on run-time speed. Qualitative tracking results for the benchmark datasets, as well as a manipulation dataset, RoboTAP \cite{vecerik2023robotap}, can be found in Figure \ref{fig:tracking_vis}. We further show qualitative results on language-guided tracking, where our model tracks objects of specific object categories that the user specifies in Figure \ref{fig:prompting_vis}.

\subsection{Multi-Object Tracking and Video Object Segmentation}
\label{exp:tracking_eval}

In each benchmark, we compare against multiple SOTA methods to see where our method stands compared to the specialized models proposed for each task.

\subsubsection{\textbf{BURST benchmark}} BURST \cite{athar2023burst} extends TAO \cite{DaveKTSR20Tao} to mask annotations. Mask annotations in BURST come from a large set of 482 classes, with the videos covering multiple types of scenes. These classes can be further divided into two subsets, the ``common" set, which contains the 78 object classes from COCO \cite{DBLP:journals/corr/LinMBHPRDZ14}, and the ``uncommon" set, which contains the remaining infrequently occurring classes from LVIS \cite{GuptaDG19LVIS}. In particular, we are interested in the ``long-tail class-guided" evaluation task, which requires us to detect and track objects corresponding to all classes, as well as predict the correct class label corresponding to each of the object tracks. This allows us to evaluate the ``open-vocabulary-ness" of \model{}, which is in stark contrast to other class-agnostic (i.e. does not require predicting object labels) existing benchmarks. We conduct our evaluations on the validation split of BURST, which contains 993 videos.

We compare the 2 baselines proposed in the BURST paper that also follows the tracking-by-detection paradigm: (1) a box tracker, which simply links per-frame detections using IoU scores followed by Hungarian matching, and (2) an STCN tracker, which uses STCN \cite{ChengTT21}, a SOTA object tracking method in the VOS literature, to propagate masks from frame $t$ to $t+1$, then uses the IoU scores of the propagated mask and the next frame detections to link object tracks.

We show quantitative results in Table \ref{tab:burst}, with results reportedly separately for all classes, the common classes, and the uncommon classes. Interestingly, despite STCN being a significantly more advanced method, the tracking quality falls behind a simple box tracker. This is due to STCN and STCN-like methods assuming ground truth object masks in the initial frame as input, and when STCN receives noisy detections, it tends to propagate the masks in an erroneous error, with the error compounding as the model tracks into the future. This effect is more noticeable on BURST as the detection task is significantly more difficult, often with parts of the object being mis-detected. \model{} achieves higher performance compared to the baselines, with a significant improvement when it comes to the tracking quality of the ``uncommon" classes.

\begin{table}
\begin{center}
\caption{\textbf{Tracking performance in BURST \cite{athar2023burst}.} Higher is better. Best performing method is \textbf{bolded}.}
\begin{tabular}{lccc}
\toprule 
Method & HOTA\textsubscript{all} & HOTA\textsubscript{com} & HOTA\textsubscript{unc} \tabularnewline
\midrule
STCN Tracker \cite{athar2023burst, ChengTT21} &5.5&17.5&2.5 \tabularnewline
Box Tracker \cite{athar2023burst} &8.2&27.0&3.6 \tabularnewline
\midrule
\model{} &\textbf{12.5}&\textbf{27.4}&\textbf{8.8}\tabularnewline
\bottomrule
\end{tabular}
\label{tab:burst}
\end{center}
\end{table}

\begin{table}
\begin{center}
\caption{\textbf{Tracking performance in UVO \cite{uvo}} Higher is better. Best performing online method is \textbf{bolded}.}
\begin{tabular}{lccc}
\toprule 
Method & $mAR100$ & $AR.5$ & $AR.75$ \tabularnewline
\midrule
\textcolor{gray}{Mask2Former VIS \cite{mask2formerVIS} (Offline)} &\textcolor{gray}{35.4}&\textcolor{gray}{-}&\textcolor{gray}{-}\tabularnewline
\midrule
MaskTrack R-CNN \cite{Yang2019vis} &17.2&-&- \tabularnewline
IDOL \cite{WuLJBYB22IDOL} &23.9&-&-\tabularnewline
TAM \cite{TAM} &24.1&-&-\tabularnewline
\midrule
\model{} &\textbf{28.1}&44.1&31.0\tabularnewline
\bottomrule
\end{tabular}
\label{tab:uvo}
\end{center}
\end{table}

\subsubsection{\textbf{UVO benchmark}} UVO \cite{uvo} is a recently proposed benchmark for Open-World Tracking. Like BURST, the mask annotations in UVO are not restricted to a small class of objects. The dataset includes annotations for anything that a human might consider as an ``object". However, the annotations are not (and cannot be) exhaustive of what an object is, since there is great ambiguity of what is a trackable entity. Since \model{} is capable of producing tracks for ``anything" that is trackable, the model often produces object tracks that may have not been annotated. Thus, we simply report the Average Recall (AR) at 0.5 and 0.75 spatio-temporal mask IOU as well as the mean Average Recall (mAR) over intervals from 0.5 to 0.95 IOU thresholds, similar to the evaluations in the COCO benchmark \cite{DBLP:journals/corr/LinMBHPRDZ14}. 

For baselines, we compare against MaskTrack R-CNN \cite{Yang2019vis}, which was proposed in the original UVO paper \cite{uvo}. We additionally consider IDOL \cite{WuLJBYB22IDOL}, a transformer-based online tracker, and TAM \cite{TAM}, a concurrent work that combines SAM with a SOTA object tracker XMem \cite{ChengS22}. Furthermore, we also include a SOTA offline tracker, Mask2Former VIS \cite{mask2formerVIS}, that extends Mask2Former for tracking through spatial-temporal attention to compare against the performance gap between offline and online methods. Note that these baselines all require training on video data, while \model{} is strictly zero-shot (i.e. not trained on any video data).

We show quantitative results in Table \ref{tab:uvo}. We can see that \model{} outperforms other online-based methods despite zero-shot. Many failure cases of \model{} are due to ground-truth object definitions and \model{} detections being misaligned, which we show in Figure \ref{fig:failure_vis}. 

\subsubsection{\textbf{DAVIS and YoutubeVOS benchmark}} We additionally conduct experiments on DAVIS'17 \cite{Pont-TusetPCASG17} and YoutubeVOS \cite{XuYFYYLPCH18}, 2 commonly used datasets for VOS tasks, where the ground truth masks are given in the initial frame, and the tracker needs to track the objects throughout the entire video. The videos are downsampled to 480p quality as per common practice in VOS literature. To reduce the amount of computation, we filter the detections in the first frame using the ground truth to reduce the amount of object tracks. In subsequent frames, we reduce the number of detections by only considering detections that have the same class labels as the detections selected in the first frame. We compare using the standard metrics: region similarity $\mathcal{J}$, contour accuracy $\mathcal{F}$, and their average $\mathcal{J\&F}$. For YoutubeVOS, we also report results separately for ``seen" or ``unseen" categories, as per conventional practice, even if \model{} does not explicitly distinguish between the two.

We consider two groups of VOS approaches as baselines: memory-based approaches such as STM \cite{OhLXK19} and STCN \cite{ChengTT21}, which use a growing history of frames and segmentation masks to propagate the segmentation in the new frame, as well as non-memory-based approaches such as SiamMask \cite{wang2019fast}, UNICORN \cite{unicorn}, Siam R-CNN \cite{VoigtlaenderLTL20}, and UNINEXT \cite{UNINEXT}, that do not use a growing history to represent object templates. All our baselines assume \textbf{ground truth} objects masks in the first frame as input, which is the standard evaluation setup of these benchmarks, while \model{} does not use any ground-truth information. Hence, we conduct additional experiments where we run the same detector (Detic \cite{zhou2022detecting}) in the initial frame to serve as input to the best-performing baseline, STCN \cite{ChengTT21}, for a more apples-to-apples comparison.

\begin{figure}[t]
\begin{center}
\includegraphics[width=\linewidth]{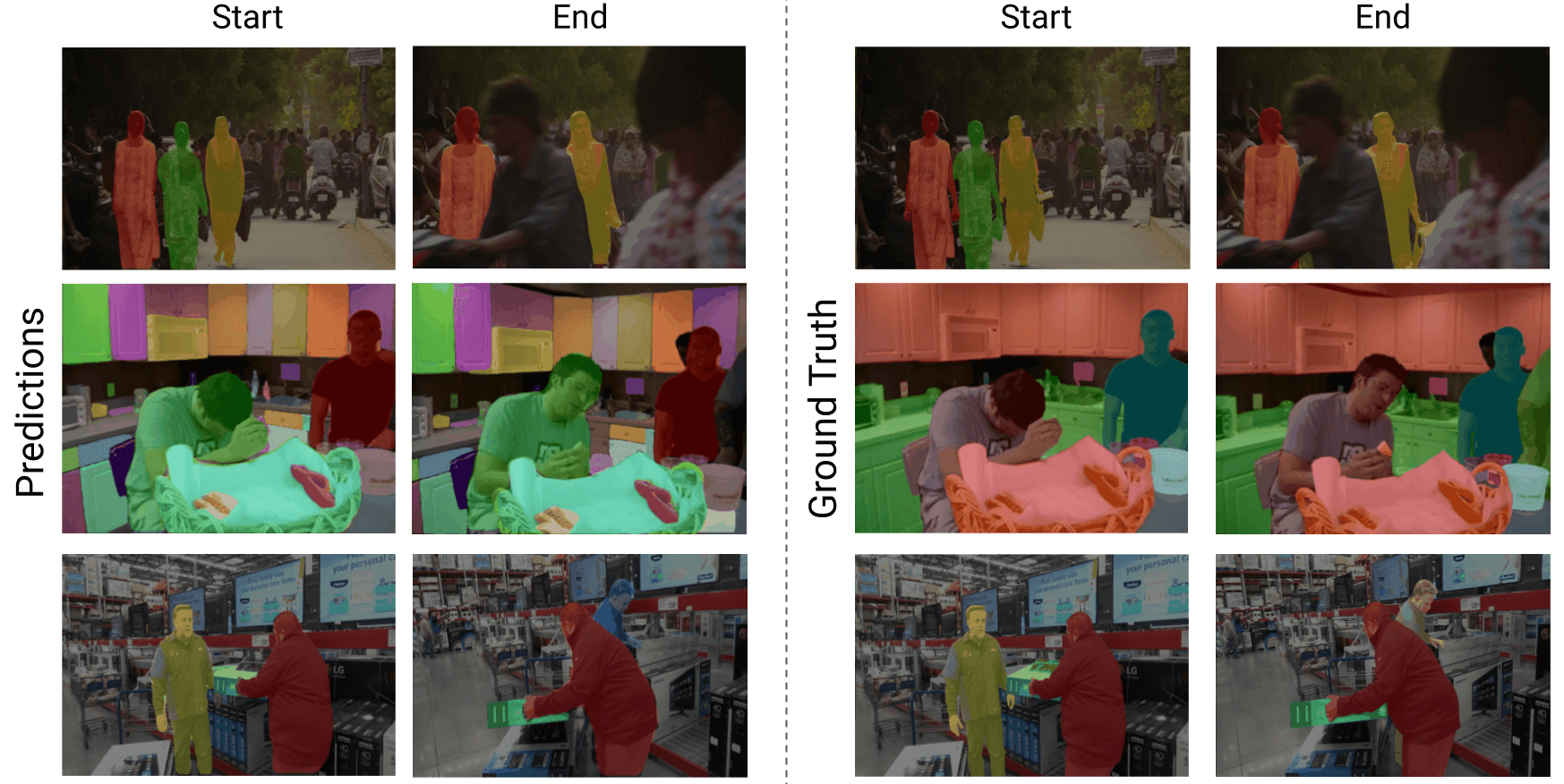}
\caption{\textbf{Failure cases of \model{}}. \textit{Top row}: The definition of classes is often ambiguous. In this case, the handbags are not considered a part of a human by our tracker. \textit{Middle row}: Even in cases without labels as in open-world settings, there can be multiple definitions of what an ``object" is. \textit{Bottom row}: Re-ID failures. In this example, we fail to match the upper half of a person (after reappearing) to an entire person (before occlusion). }
  \label{fig:failure_vis}
\end{center}
\end{figure}

We show quantitative results for our model and baselines on YoutubeVOS and DAVIS in Table \ref{tab:vos}. We can see that the performance drops considerably when we switch from GT mask inputs to detection masks for STCN. This is due to multiple reasons: (1) ambiguities in ``what" to track, as seen in Figure \ref{fig:failure_vis}, where the tracker is missing the handheld bags of the ladies, and (2) existing VOS-based approaches starts to propagate the errors from detection predictions to future frames, which we also observed in BURST. This difference is further exaggerated in YoutubeVOS, where the detection problem is harder: if the detector fails to detect the object instance on the annotated frame, it's not possible to associate the object track with the correct instance ID in the GT, even if the detections are perfect in all subsequent frames. Since annotations are usually supplied in the first frame an object appears, there are many cases where the annotated frame is not an easy frame to detect the object, as the entities might have just started entering the scene.
\begin{table*}
\begin{center}
\caption{\textbf{Performance on YoutubeVOS 2018 \cite{XuYFYYLPCH18} and DAVIS'17 \cite{Pont-TusetPCASG17} .} Higher is better.}
\begin{tabular}{llcccccccc}
\toprule 
\multirow{2}{*}{\ \ \ \ }&\multirow{2}{*}{Method} & \multicolumn{5}{c}{YoutubeVOS 2018 Val}  & \multicolumn{3}{c}{DAVIS'17 Val} \\
\cmidrule(lr){3-7} \cmidrule(lr){8-10}
& & $\mathcal{G}$ & $\mathcal{J}_s$ & $\mathcal{F}_s$ & $\mathcal{J}_u$ & $\mathcal{F}_u$ & $\mathcal{J\&F}$ & $\mathcal{J}$ & $\mathcal{F}$\\

\midrule
\multirow{6}{*}{\textbf{Ground-Truth}}
&SiamMask~\cite{wang2019fast}&52.8&60.2&58.2&45.1&47.7&56.4&54.3&58.5\tabularnewline
\multirow{6}{*}{\textbf{masks} at $t=0$}
&Unicorn~\cite{unicorn}&-&-&-&-&-&69.2&65.2&73.2\tabularnewline
&Siam R-CNN~\cite{VoigtlaenderLTL20}&73.2&73.5&-&66.2&-&70.6&66.1&75.0\tabularnewline
&UNINEXT~\cite{UNINEXT}& 78.6 & 79.9 & 84.9 & 70.6 & 79.2 & 81.8 & 77.7 & 85.8\tabularnewline
&STM~\cite{OhLXK19}&79.4&79.7&84.2&72.8&80.9&81.8&79.2&84.3\tabularnewline
&STCN~\cite{ChengTT21}&83.0&81.9&86.5&77.9&85.7&85.4&82.2&88.6\tabularnewline
\midrule
\multirow{1}{*}{\textbf{Detected masks}}
& Detic + STCN &58.8&68.1&71.7&45.3&50.3&76.5&74.5&78.5\tabularnewline
\multirow{1}{*}{at $t=0$}
& \model{} &62.2&65.9&69.4&53.7&59.8&74.8&73.2&76.4\tabularnewline
\bottomrule
\end{tabular}
\label{tab:vos}
\end{center}
\end{table*}
\begin{figure}[!ht]
\begin{center}
\includegraphics[width=\linewidth]{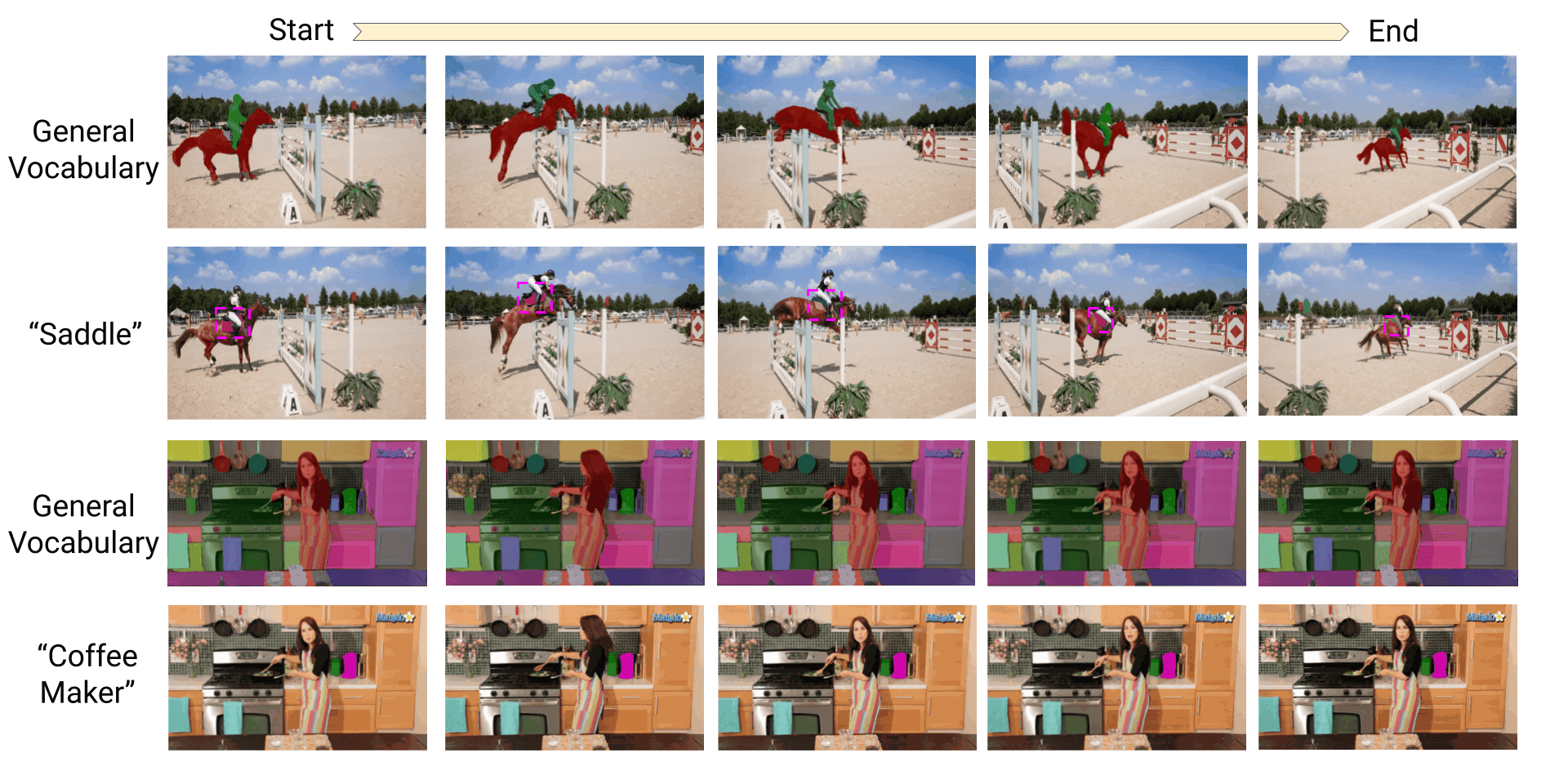}
\caption{\textbf{Language-guided tracking.} \textit{Top two rows}: The user prompts with ``saddle'', and the model can recover the missing tracks (colored pink). We also draw a bounding box for clearer visibility. \textit{Bottom two rows}: The same interface can be easily applied if the user only wishes to track specific objects in a video like ``coffee makers" (colored pink and green).}
  \label{fig:prompting_vis}
\end{center}
\end{figure}

\subsection{Promptable Open-Vocabulary Tracking}
The object tracks our method delivers are associated with corresponding category labels. A user can specify a category to track, and then our method can deliver object tracks only of that specific category. We show such an example in Figure \ref{fig:prompting_vis}. When the human prompts the tracker for a specific object category, we use a lower confidence threshold $\lambda_c$ for the detector only for instances of the mentioned category, as the human provides a powerful cue that the mentioned object is in the video. We show more examples in the supplementary file, as well as tracking objects based on open-world referential utterances by using a referential detector \cite{liu2023grounding}. 

\begin{table}
\begin{center}
\caption{\textbf{Average running time} in milliseconds for the individual components in \model{}.}
\begin{tabular}{lc}
\toprule 
Component & Avg. Run Time (ms) \tabularnewline
\midrule
Detic (with TTA) & 1350 \tabularnewline
Optical Flow & 275 \tabularnewline
Box Warping & 150 \tabularnewline
SAM & 550 \tabularnewline
Spawning/ReID & 100 \tabularnewline
\bottomrule
\end{tabular}
\label{tab:exec_time}
\end{center}
\end{table}

\subsection{Ablations}
We ablate design choices of our method in Table \ref{tab:ablation} in the DAVIS benchmark. Not using segmentation to adapt the object boxes gives a much worse performance, which shows that segmentation helps to track by providing higher quality boxes, as opposed to computing it simply as a post-processing step. Not using mask cycle consistency hurts by a small amount. Skipping the bounding box refinement also hurts by a great amount, which indicates that an optical flow-based motion model alone is not sufficient to obtain good box prompts for the segmenter. Skipping the flow-based motion propagation also unsurprisingly hurts performance, since the bounding box refinement alone cannot recover the correct object box, especially for frames where the motion is large. 

\begin{table}
\centering{}
\caption{\textbf{Ablative analysis in DAVIS'17}. We show the relative changes w.r.t. the complete model (top row). }
\begin{tabular}{lcccc}
\toprule
\multirow{2}{*}{Method} & \multicolumn{3}{c}{DAVIS'17 Val} \\
\cmidrule(lr){2-4} & $\mathcal{J\&F}$ & $\mathcal{J}$ & $\mathcal{F}$\\
\midrule
\model{} &74.8&73.2&76.4 \tabularnewline
\midrule
$-$ Box Adaptation after Segmentation &  
 -3.1&-2.9&-3.3\tabularnewline
$-$ Motion Based Propagation &  
 -4.3&-3.9&-4.5 \tabularnewline
$-$ Box Refinement &  
 -3.7&-3.4&-3.9 \tabularnewline
$-$ Mask Cycle Consistency &  
 -2.0&-1.4&-2.7 \tabularnewline
\bottomrule
\end{tabular}
\label{tab:ablation}
\end{table}

\subsection{Running Time Analysis}
We analyze the running time of the individual components in \model{} in Table \ref{tab:exec_time}. The results are reported over the average of all the frames in the videos. The model runs at 0.41 FPS on an Nvidia V100 and costs around 18GB of VRAM to run on 480p videos, without caching anything to disk. We can see that most of the running time is spent in the Detic detector, with the SAM segmenter coming in second. Test time augmentations (TTA) incur a big overhead for Detic, and for scenes where detection is easy, noticeable speedups can be achieved by turning off TTA.

\section{Limitations and Future Work}
\label{limitations}

A limitation of \model{} is the use of pre-trained features for feature matching for re-identifying object tracks. Empirically we observed that these features are not necessarily temporally consistent when used directly out-of-the-box, leading to many mistakes in the Re-ID process. We will explore the possibility of training more general-purpose re-id networks as an extension in our future work. 
Augmenting SAM with extra modules \cite{sam_hq} for tighter spatial-temporal reasoning, where mask query tokens attend to previous frames to be better conditioned on a track's history is another interesting avenue of future work.

\section{Conclusion}
\label{conclusions}
We present \model{}, a zero-shot framework for open-vocabulary visual tracking, that re-purposes modules of large-scale pre-trained models for object tracking in videos without any training or finetuning. Our model can be applied equally well for video object segmentation and multi-object tracking across various benchmarks and helps unify the video tracking and segmentation literature, which has been oven-fragmented using different evaluation protocols and ground-truth information at test time. Instead of specifying what to track with ground-truth object masks, which are hard to annotate and provide, our tracking framework offers a language interface over what to focus on and track in videos. Thanks to its simplicity, we hope that our model can serve as an upgradeable and extendable baseline for future work in open-world and open-vocabulary video tracking literature.


\bibliographystyle{IEEEtran}
\bibliography{IEEEabrv,99_refs}

\begin{thebibliography}{10}
\providecommand{\url}[1]{#1}
\csname url@rmstyle\endcsname
\providecommand{\newblock}{\relax}
\providecommand{\bibinfo}[2]{#2}
\providecommand\BIBentrySTDinterwordspacing{\spaceskip=0pt\relax}
\providecommand\BIBentryALTinterwordstretchfactor{4}
\providecommand\BIBentryALTinterwordspacing{\spaceskip=\fontdimen2\font plus
\BIBentryALTinterwordstretchfactor\fontdimen3\font minus
  \fontdimen4\font\relax}
\providecommand\BIBforeignlanguage[2]{{%
\expandafter\ifx\csname l@#1\endcsname\relax
\typeout{** WARNING: IEEEtran.bst: No hyphenation pattern has been}%
\typeout{** loaded for the language `#1'. Using the pattern for}%
\typeout{** the default language instead.}%
\else
\language=\csname l@#1\endcsname
\fi
#2}}

\bibitem{citeulike:3504612}
M.~Andriluka, S.~Roth, and B.~Schiele, ``People-tracking-by-detection and
  people-detection-by-tracking,'' \emph{CVPR}, 2008.

\bibitem{baseline1}
X.~Weng and K.~Kitani, ``A baseline for 3d multi-object tracking,''
  \emph{arXiV}, 2019.

\bibitem{zhou2022detecting}
X.~Zhou, R.~Girdhar, A.~Joulin, P.~Kr{\"a}henb{\"u}hl, and I.~Misra,
  ``Detecting twenty-thousand classes using image-level supervision,'' in
  \emph{ECCV}, 2022.

\bibitem{kirillov2023segment}
A.~Kirillov, E.~Mintun, N.~Ravi, H.~Mao, C.~Rolland, L.~Gustafson, T.~Xiao,
  S.~Whitehead, A.~C. Berg, W.-Y. Lo, \emph{et~al.}, ``Segment anything,''
  \emph{arXiv}, 2023.

\bibitem{GMFlow}
H.~Xu, J.~Zhang, J.~Cai, H.~Rezatofighi, and D.~Tao, ``Gmflow: Learning optical
  flow via global matching,'' in \emph{CVPR}, 2022.

\bibitem{bergmann2019tracking}
P.~Bergmann, T.~Meinhardt, and L.~Leal-Taixe, ``Tracking without bells and
  whistles,'' in \emph{CVPR}, 2019.

\bibitem{Pont-TusetPCASG17}
J.~Pont{-}Tuset, F.~Perazzi, S.~Caelles, P.~Arbelaez, A.~Sorkine{-}Hornung, and
  L.~V. Gool, ``The 2017 {DAVIS} challenge on video object segmentation,''
  \emph{arXiV}, 2017.

\bibitem{XuYFYYLPCH18}
N.~Xu, L.~Yang, Y.~Fan, J.~Yang, D.~Yue, Y.~Liang, B.~L. Price, S.~Cohen, and
  T.~S. Huang, ``Youtube-vos: Sequence-to-sequence video object segmentation,''
  in \emph{ECCV}, 2018.

\bibitem{uvo}
W.~Wang, M.~Feiszli, H.~Wang, and D.~Tran, ``Unidentified video objects: {A}
  benchmark for dense, open-world segmentation,'' \emph{CoRR}, 2021.

\bibitem{athar2023burst}
A.~Athar, J.~Luiten, P.~Voigtlaender, T.~Khurana, A.~Dave, B.~Leibe, and
  D.~Ramanan, ``Burst: A benchmark for unifying object recognition,
  segmentation and tracking in video,'' in \emph{WACV}, 2023.

\bibitem{vecerik2023robotap}
M.~Vecerik, C.~Doersch, Y.~Yang, T.~Davchev, Y.~Aytar, G.~Zhou, R.~Hadsell,
  L.~Agapito, and J.~Scholz, ``Robotap: Tracking arbitrary points for few-shot
  visual imitation,'' \emph{arXiv}, 2023.

\bibitem{tbd}
M.~Breitenstein, F.~Reichlin, B.~Leibe, E.~Koller-Meier, and L.~Van~Gool,
  ``Online multi-person tracking-by-detection from a single, uncalibrated
  camera.'' \emph{IEEE Transactions on Pattern Analysis and Machine
  Intelligence}, 2011.

\bibitem{FasterRCNN}
S.~Ren, K.~He, R.~Girshick, and J.~Sun, ``{Faster R--CNN}: Towards real-time
  object detection with region proposal networks,'' in \emph{NeurIPS}, 2015.

\bibitem{trackformer}
T.~Meinhardt, A.~Kirillov, L.~Leal{-}Taix{\'{e}}, and C.~Feichtenhofer,
  ``Trackformer: Multi-object tracking with transformers,'' \emph{CoRR}, 2021.

\bibitem{athar2022hodor}
A.~Athar, J.~Luiten, A.~Hermans, D.~Ramanan, and B.~Leibe, ``Hodor: High-level
  object descriptors for object re-segmentation in video learned from static
  images,'' in \emph{CVPR}, 2022.

\bibitem{DBLP:journals/corr/abs-2012-15460}
P.~Sun, Y.~Jiang, R.~Zhang, E.~Xie, J.~Cao, X.~Hu, T.~Kong, Z.~Yuan, C.~Wang,
  and P.~Luo, ``Transtrack: Multiple-object tracking with transformer,''
  \emph{CoRR}, 2020.

\bibitem{liu2023grounding}
S.~Liu, Z.~Zeng, T.~Ren, F.~Li, H.~Zhang, J.~Yang, C.~Li, J.~Yang, H.~Su,
  J.~Zhu, and L.~Zhang, ``Grounding dino: Marrying dino with grounded
  pre-training for open-set object detection,'' 2023.

\bibitem{zhang2022glipv2}
H.~Zhang, P.~Zhang, X.~Hu, Y.-C. Chen, L.~H. Li, X.~Dai, L.~Wang, L.~Yuan,
  J.-N. Hwang, and J.~Gao, ``Glipv2: Unifying localization and vision-language
  understanding,'' 2022.

\bibitem{zhou2020tracking}
X.~Zhou, V.~Koltun, and P.~Kr{\"a}henb{\"u}hl, ``Tracking objects as points,''
  in \emph{ECCV}, 2020.

\bibitem{bewley2016simple}
A.~Bewley, Z.~Ge, L.~Ott, F.~Ramos, and B.~Upcroft, ``Simple online and
  realtime tracking,'' in \emph{ICIP}, 2016.

\bibitem{wojke2017simple}
N.~Wojke, A.~Bewley, and D.~Paulus, ``Simple online and realtime tracking with
  a deep association metric,'' in \emph{ICIP}, 2017.

\bibitem{tang2017multiple}
S.~Tang, M.~Andriluka, B.~Andres, and B.~Schiele, ``Multiple people tracking by
  lifted multicut and person re-identification,'' in \emph{CVPR}, 2017.

\bibitem{zhang2021fairmot}
Y.~Zhang, C.~Wang, X.~Wang, W.~Zeng, and W.~Liu, ``Fairmot: On the fairness of
  detection and re-identification in multiple object tracking,'' \emph{IJCV},
  2021.

\bibitem{radford2021learning}
A.~Radford, J.~W. Kim, C.~Hallacy, A.~Ramesh, G.~Goh, S.~Agarwal, G.~Sastry,
  A.~Askell, P.~Mishkin, J.~Clark, \emph{et~al.}, ``Learning transferable
  visual models from natural language supervision,'' in \emph{International
  conference on machine learning}.\hskip 1em plus 0.5em minus 0.4em\relax PMLR,
  2021, pp. 8748--8763.

\bibitem{jia2021scaling}
C.~Jia, Y.~Yang, Y.~Xia, Y.-T. Chen, Z.~Parekh, H.~Pham, Q.~Le, Y.-H. Sung,
  Z.~Li, and T.~Duerig, ``Scaling up visual and vision-language representation
  learning with noisy text supervision,'' in \emph{ICML}, 2021.

\bibitem{devlin2018bert}
J.~Devlin, M.-W. Chang, K.~Lee, and K.~N. Toutanova, ``Bert: Pre-training of
  deep bidirectional transformers for language understanding,'' in
  \emph{NAACL}, 2019.

\bibitem{pennington2014glove}
J.~Pennington, R.~Socher, and C.~D. Manning, ``Glove: Global vectors for word
  representation,'' in \emph{EMNLP}, 2014.

\bibitem{zareian2021open}
A.~Zareian, K.~D. Rosa, D.~H. Hu, and S.-F. Chang, ``Open-vocabulary object
  detection using captions,'' in \emph{CVPR}, 2021.

\bibitem{bansal2018zero}
A.~Bansal, K.~Sikka, G.~Sharma, R.~Chellappa, and A.~Divakaran, ``Zero-shot
  object detection,'' in \emph{ECCV}, 2018.

\bibitem{rahman2020improved}
S.~Rahman, S.~Khan, and N.~Barnes, ``Improved visual-semantic alignment for
  zero-shot object detection,'' in \emph{AAAI}, 2020.

\bibitem{gu2021open}
X.~Gu, T.-Y. Lin, W.~Kuo, and Y.~Cui, ``Open-vocabulary object detection via
  vision and language knowledge distillation,'' \emph{ICLR}, 2022.

\bibitem{minderer2022simple}
M.~Minderer, A.~Gritsenko, A.~Stone, M.~Neumann, D.~Weissenborn,
  A.~Dosovitskiy, A.~Mahendran, A.~Arnab, M.~Dehghani, Z.~Shen, \emph{et~al.},
  ``Simple open-vocabulary object detection with vision transformers,''
  \emph{arXiv}, 2022.

\bibitem{shi1998motion}
J.~Shi and J.~Malik, ``Motion segmentation and tracking using normalized
  cuts,'' in \emph{ICCV}, 1998.

\bibitem{grundmann2010efficient}
M.~Grundmann, V.~Kwatra, M.~Han, and I.~Essa, ``Efficient hierarchical
  graph-based video segmentation,'' in \emph{CVPR}.\hskip 1em plus 0.5em minus
  0.4em\relax IEEE, 2010.

\bibitem{ochs2013segmentation}
P.~Ochs, J.~Malik, and T.~Brox, ``Segmentation of moving objects by long term
  video analysis,'' \emph{TPAMI}, vol.~36, no.~6, 2013.

\bibitem{Fragkiadaki_2015_CVPR}
K.~Fragkiadaki, P.~Arbelaez, P.~Felsen, and J.~Malik, ``Learning to segment
  moving objects in videos,'' in \emph{CVPR}, 2015.

\bibitem{bideau2018best}
P.~Bideau, A.~RoyChowdhury, R.~R. Menon, and E.~Learned-Miller, ``The best of
  both worlds: Combining cnns and geometric constraints for hierarchical motion
  segmentation,'' in \emph{CVPR}, 2018.

\bibitem{liu2022opening}
Y.~Liu, I.~E. Zulfikar, J.~Luiten, A.~Dave, D.~Ramanan, B.~Leibe,
  A.~O{\v{s}}ep, and L.~Leal-Taix{\'e}, ``Opening up open world tracking,'' in
  \emph{CVPR}, 2022.

\bibitem{li2022tracking}
S.~Li, M.~Danelljan, H.~Ding, T.~E. Huang, and F.~Yu, ``Tracking every thing in
  the wild,'' in \emph{ECCV}, 2022.

\bibitem{ovsep2018track}
A.~O{\v{s}}ep, W.~Mehner, P.~Voigtlaender, and B.~Leibe, ``Track, then decide:
  Category-agnostic vision-based multi-object tracking,'' in \emph{ICRA}, 2018.

\bibitem{LiuL00W0LG21}
Z.~Liu, Y.~Lin, Y.~Cao, H.~Hu, Y.~Wei, Z.~Zhang, S.~Lin, and B.~Guo, ``Swin
  transformer: Hierarchical vision transformer using shifted windows,'' in
  \emph{ICCV}, 2021.

\bibitem{VaswaniSPUJGKP17}
A.~Vaswani, N.~Shazeer, N.~Parmar, J.~Uszkoreit, L.~Jones, A.~N. Gomez,
  L.~Kaiser, and I.~Polosukhin, ``Attention is all you need,'' in
  \emph{NeurIPS}, 2017.

\bibitem{pan2009recurrent}
P.~Pan, F.~Porikli, and D.~Schonfeld, ``Recurrent tracking using multifold
  consistency,'' in \emph{Proceedings of the Eleventh IEEE International
  Workshop on Performance Evaluation of Tracking and Surveillance}, vol.~3,
  2009.

\bibitem{conf/eccv/SundaramBK10}
N.~Sundaram, T.~Brox, and K.~Keutzer, ``Dense point trajectories by
  gpu-accelerated large displacement optical flow.'' in \emph{ECCV}, 2010.

\bibitem{DaveKTSR20Tao}
A.~Dave, T.~Khurana, P.~Tokmakov, C.~Schmid, and D.~Ramanan, ``{TAO:} {A}
  large-scale benchmark for tracking any object,'' in \emph{ECCV}, 2020.

\bibitem{DBLP:journals/corr/LinMBHPRDZ14}
\BIBentryALTinterwordspacing
T.~Lin, M.~Maire, S.~J. Belongie, L.~D. Bourdev, R.~B. Girshick, J.~Hays,
  P.~Perona, D.~Ramanan, P.~Doll{\'{a}}r, and C.~L. Zitnick, ``Microsoft
  {COCO:} common objects in context,'' \emph{CoRR}, vol. abs/1405.0312, 2014.
  [Online]. Available: \url{http://arxiv.org/abs/1405.0312}
\BIBentrySTDinterwordspacing

\bibitem{GuptaDG19LVIS}
A.~Gupta, P.~Doll{\'{a}}r, and R.~B. Girshick, ``{LVIS:} {A} dataset for large
  vocabulary instance segmentation,'' in \emph{CVPR}, 2019.

\bibitem{ChengTT21}
H.~K. Cheng, Y.~Tai, and C.~Tang, ``Rethinking space-time networks with
  improved memory coverage for efficient video object segmentation,'' in
  \emph{NeurIPS}, 2021.

\bibitem{mask2formerVIS}
B.~Cheng, A.~Choudhuri, I.~Misra, A.~Kirillov, R.~Girdhar, and A.~G. Schwing,
  ``Mask2former for video instance segmentation,'' \emph{CoRR}, 2021.

\bibitem{Yang2019vis}
L.~Yang, Y.~Fan, and N.~Xu, ``Video instance segmentation,'' \emph{CoRR}, 2019.

\bibitem{WuLJBYB22IDOL}
J.~Wu, Q.~Liu, Y.~Jiang, S.~Bai, A.~L. Yuille, and X.~Bai, ``In defense of
  online models for video instance segmentation,'' in \emph{ECCV}, 2022.

\bibitem{TAM}
J.~Yang, M.~Gao, Z.~Li, S.~Gao, F.~Wang, and F.~Zheng, ``Track anything:
  Segment anything meets videos,'' \emph{CoRR}, 2023.

\bibitem{ChengS22}
H.~K. Cheng and A.~G. Schwing, ``Xmem: Long-term video object segmentation with
  an atkinson-shiffrin memory model,'' in \emph{ECCV}, 2022.

\bibitem{OhLXK19}
S.~W. Oh, J.~Lee, N.~Xu, and S.~J. Kim, ``Video object segmentation using
  space-time memory networks,'' in \emph{ICCV}, 2019.

\bibitem{wang2019fast}
Q.~Wang, L.~Zhang, L.~Bertinetto, W.~Hu, and P.~H. Torr, ``Fast online object
  tracking and segmentation: A unifying approach,'' in \emph{CVPR}, 2019.

\bibitem{unicorn}
B.~Yan, Y.~Jiang, P.~Sun, D.~Wang, Z.~Yuan, P.~Luo, and H.~Lu, ``Towards grand
  unification of object tracking,'' in \emph{ECCV}, 2022.

\bibitem{VoigtlaenderLTL20}
P.~Voigtlaender, J.~Luiten, P.~H.~S. Torr, and B.~Leibe, ``Siam {R-CNN:} visual
  tracking by re-detection,'' in \emph{CVPR}, 2020.

\bibitem{UNINEXT}
B.~Yan, Y.~Jiang, J.~Wu, D.~Wang, Z.~Yuan, P.~Luo, and H.~Lu, ``Universal
  instance perception as object discovery and retrieval,'' in \emph{CVPR},
  2023.

\bibitem{sam_hq}
L.~Ke, M.~Ye, M.~Danelljan, Y.~Liu, Y.-W. Tai, C.-K. Tang, and F.~Yu, ``Segment
  anything in high quality,'' \emph{arXiv}, 2023.

\end{thebibliography}

\end{document}